\documentclass{ieeeaccess}
\usepackage{cite}
\usepackage{amsmath,amssymb,amsfonts}
\usepackage{graphicx}
\usepackage{textcomp}
\usepackage[small]{caption}
\usepackage{graphicx}
\usepackage{amsmath}
\usepackage{amssymb}
\usepackage{booktabs}
\usepackage[noend]{algpseudocode}
\usepackage{multirow}
\usepackage{algorithm}
\usepackage{graphics}

\makeatletter
\AtBeginDocument{\DeclareMathVersion{bold}
\SetSymbolFont{operators}{bold}{T1}{times}{b}{n}
\SetSymbolFont{NewLetters}{bold}{T1}{times}{b}{it}
\SetMathAlphabet{\mathrm}{bold}{T1}{times}{b}{n}
\SetMathAlphabet{\mathit}{bold}{T1}{times}{b}{it}
\SetMathAlphabet{\mathbf}{bold}{T1}{times}{b}{n}
\SetMathAlphabet{\mathtt}{bold}{OT1}{pcr}{b}{n}
\SetSymbolFont{symbols}{bold}{OMS}{cmsy}{b}{n}
\renewcommand\boldmath{\@nomath\boldmath\mathversion{bold}}}
\makeatother

\def\BibTeX{{\rm B\kern-.05em{\sc i\kern-.025em b}\kern-.08em
    T\kern-.1667em\lower.7ex\hbox{E}\kern-.125emX}}

\begin{document}
\history{This work is under review by the IEEE ACCESS journal.}
\doi{}
\title{Safe Policy Exploration Improvement via Subgoals}
\author{\uppercase{Brian Angulo}\authorrefmark{1, 4},
\uppercase{Gregory Gorbov}\authorrefmark{1}, 
\uppercase{Aleksandr Panov}\authorrefmark{1,2,3} and 
\uppercase{Konstantin Yakovlev}\authorrefmark{2,3}}

\address[1]{Moscow Institute of Physics and Technology, 141701 Dolgoprudny, Russia}
\address[2]{Federal Research Center “Computer Science and Control” of the Russian Academy of Sciences, 119333 Moscow, Russia}
\address[3]{AIRI, 123112 Moscow, Russia}
\address[4]{LLC “Integrant”, 127204 Moscow, Russia}
\tfootnote{This work was partially supported by the Analytical Center for the Government of the Russian Federation in accordance with the subsidy agreement (agreement identifier 000000D730321P5Q0002; grant  No. 70-2021-00138).}


\markboth
{Author \headeretal: Preparation of Papers for IEEE TRANSACTIONS and JOURNALS}
{Author \headeretal: Preparation of Papers for IEEE TRANSACTIONS and JOURNALS}

\corresp{Corresponding author: Brian Angulo (e-mail: brianangulo.yauri@gmail.com).}

\begin{abstract}

Reinforcement learning is a widely used approach to autonomous navigation, showing potential in various tasks and robotic setups. Still, it often struggles to reach distant goals when safety constraints are imposed (e.g., the wheeled robot is prohibited from moving close to the obstacles). One of the main reasons for poor performance in such setups, which is common in practice, is that the need to respect the safety constraints degrades the exploration capabilities of an RL agent. To this end, we introduce a novel learnable algorithm that is based on decomposing the initial problem into smaller sub-problems via intermediate goals, on the one hand, and respects the limit of the cumulative safety constraints, on the other hand -- SPEIS(Safe Policy Exploration Improvement via Subgoals). It comprises the two coupled policies trained end-to-end: subgoal and safe. The subgoal policy is trained to generate the subgoal based on the transitions from the buffer of the safe (main) policy that helps the safe policy to reach distant goals. Simultaneously, the safe policy maximizes its rewards while attempting not to violate the limit of the cumulative safety constraints, thus providing a certain level of safety. We evaluate SPEIS in a wide range of challenging (simulated) environments that involve different types of robots in two different environments: autonomous vehicles from the POLAMP environment and car, point, doggo, and sweep from the safety-gym environment. We demonstrate that our method consistently outperforms state-of-the-art competitors and can significantly reduce the collision rate while maintaining high success rates (higher by 80\% compared to the best-performing methods).

\end{abstract}


\titlepgskip=-21pt

\maketitle

\section{Introduction}

Deep Reinforcement Learning has demonstrated tremendous success in many high-dimensional control problems, including autonomous navigation~\cite{zhu2021survey}. Within RL, the interaction of the agent with the environment is modeled as a Markov decision process (MDP)~\cite{sutton2018reinforcement}, where the goal is to optimize the expected cumulative reward. While reinforcement learning algorithms have had great success in the field of autonomous navigation \cite{aradi2020survey}, they cannot be straightforwardly applied to real autonomous systems because these algorithms have significant freedom to explore any behavior that could improve its performance, including those that might cause damage \cite{ha2020learning}. To this end, it is crucial to take into account safety constraints during the training to decrease unsafe behaviors of these algorithms in real autonomous systems. A well-known approach to consider safety constraints in RL is a Constrained Markov Decision Process (CMDP)~\cite{altman1999constrained}. In this paper, we will address CMDP for the navigation problem to ensure the safe behavior of an agent. 

\begin{figure}[t]
    \centering
    \includegraphics[width=0.48\textwidth]{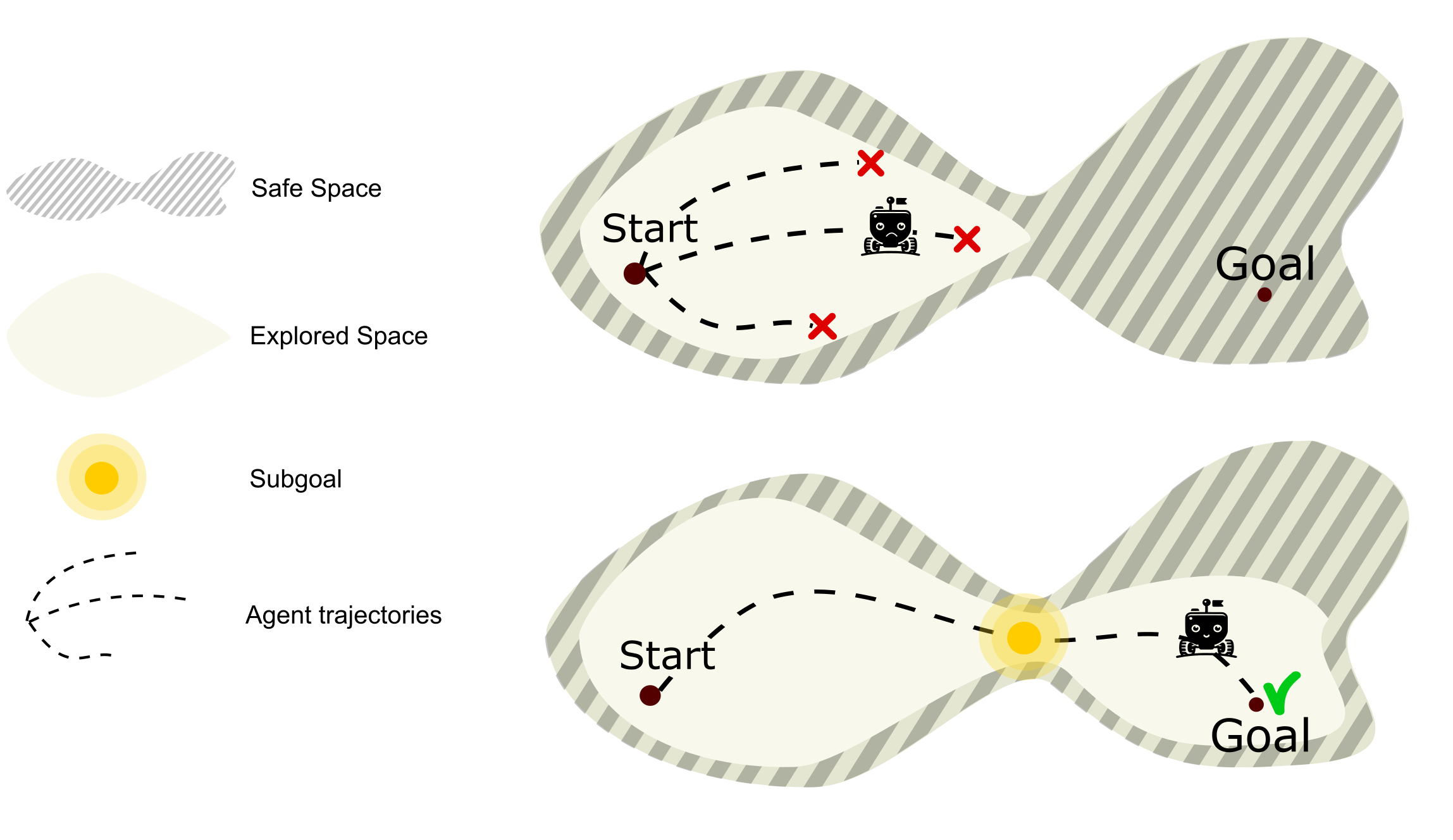}
    \caption{Safe exploration by utilizing the subgoals. Top: A safe policy often lacks sufficient exploration to effectively solve long-horizon tasks (i.e., when an agent needs to execute a long sequence of actions to reach the goal). Bottom: To mitigate this problem, we propose to generate subgoals inside the safe region to increase the exploration capabilities and, therefore, to increase the chance of successfully reaching the distant goal.}
\label{fig:vis-abstract}
\end{figure}

One of the most widely used approaches for solving CMDP is the Lagrangian method~\cite{chow2017risk}. It reduces the CMDP to an unconstrained problem via Lagrange multipliers (safety weights). The safety weights are used to assign a value to the violation of the safety constraints during training and can be optimized based on how much we violate the limit of the cumulative safety constraints. However, considering safety constraints during the training significantly reduces the agent's exploration -- see Fig.~\ref{fig:vis-abstract}. This problem with the lack of exploration is well-known as safe exploration in the field of safe reinforcement learning and is highlighted in~\cite{ray2019benchmarking, jayant2022model}. 

In addition to safe exploration, solving practical navigation problems often requires reaching distant goals in continuous domains with high-dimensional observations. Because of that, the agent struggles to perform temporally extended reasoning: the well-known long-horizon problem~\cite{sutton1999between}. 
One of the common approaches to address the latter problem is to utilize a hierarchical approach that decomposes the initial (long horizon) problem into smaller (small horizon) sub-problems via intermediate goals (subgoals)~\cite{nachum2018, chane2021goal} that help to guide the exploration.

In this work, we suggest SPEIS, an end-to-end learnable method that utilizes a hierarchical approach for solving CMDP. This algorithm comprises two reinforcement learning approaches (safe and hierarchical) to find a policy that solves long-horizon navigation tasks considering the limit of the cumulative safety constraints. The subgoal policy aims to guide the robot to reach the distant goals, and it is trained based on the experience of the safe policy. Simultaneously, we train the safe policy to find a trade-off between not violating the limit of cumulative safety constraints and maximizing the cumulative reward while relying on a subgoal to achieve distant goals. As a result, SPEIS acquires a safe behavior that allows the agent to estimate whether it is safer to stay in place than move forward and consequently violate the limit of cumulative safety constraints or collide with static obstacles. This property, along with the subgoal policy, allows it to greatly reduce the violation of safety constraints and to deliver a low collision rate, 3\% on average for all considered robots, while providing a comparable success rate, above 80\% on average, to that of the current best performing methods.


\section{Related Work}

Recently, the issue of navigation with safety constraints has garnered increased interest among researchers, driven by integrating RL methods into real-world systems. Despite this growing attention, methods for long-term navigation tasks under safety constraints remain a persistent challenge.

\subsection{Safe Reinforcement Learning}
Within the domain of safe reinforcement learning, two primary approaches are distinguished~\cite{gu2022review, liu2021policy}: methods that integrate a safety component into the actor's objective function and methods that replace each action of the agent with a safe alternative.

\subsubsection{Safety component to actor objective}
Among the methods that consider an additional objective to address safety constraints, it is worth mentioning the SAC Lagrangian~\cite{yang2021wcsac} and NeuralConditionedSAC~\cite{huang2021risk} methods, which are based on SAC and train an additional critic that predicts the cumulative safety signal. Consequently, during updates, the policy takes into account the sum of the values of the two critics.

\subsubsection{Substitute action to safe action} In general, such approaches involve a controller that outputs an unsafe (greedy) action. A Quadratic Programming (QP) problem is formulated to identify a safe action, where the objective function represents the Euclidean distance between the new safe action and the greedy one. The methods differ primarily in the formulation of linear constraints for this problem. For instance, in SPICE~\cite{anderson2022guiding}, the function defining whether a state is dangerous or not was represented as a linear function. In this case, it was demonstrated how linear constraints for the QP problem could be analytically derived using a Taylor approximation for the world model. Another example is SafetyLayer\cite{dalal2018safe}, where the additional model is pre-trained to predict the cost signal associated with each action using offline data. Using this model, a quadratic programming optimization problem is formulated via Taylor approximation for the cost function and solved to replace each action with a safe one. The main drawback of such approaches is that during algorithm testing, a significant amount of time is spent solving the QP problem for each action.

\subsection{Hierarchical Reinforcement Learning}
Hierarchical approaches in RL are typically employed to enhance the exploration of policy or when an agent needs to take a large number of steps to achieve a goal. In the latter case, intermediate sub-goals aid in breaking down the task into simpler ones\cite{hutsebaut2022hierarchical}. Some approaches involve generating sub-goals that the policy must achieve within a limited number of steps, as seen in methods like HAC\cite{levy2017hierarchical}, HiRO\cite{nachum2018data}, HRAC\cite{zhang2020generating}. In these cases, two neural networks are used during inference. Other RIS\cite{chane2021goal} utilize sub-goals solely as an additional component for the policy's objective function, where only the policy is used during inference.

\subsection{Hierarchical and Safety Reinforcement Learning}
There are very few studies on integrating hierarchical approaches with safety constraints. Among these, LyapunovRRT~\cite{xiong2022model} stands out as an example. It is proposed that the RRT* planner be employed for subgoal generation and the co-learning TD3 policy with Neural Lyapunov Function be employed to generate ROA (region of attraction) on the planner's path to ensure safety. It is also worth mentioning the method proposed in~\cite{roza2022safe}, which integrates SafetyLayer and HiRO. Our approach differs in that we propose using the safety module only during training, so during inference, time is only spent computing an action already aware of safety constraints. In contrast, in SafetyLayer+HiRO, an additional optimization task is solved for each action to replace it with a safe one. Furthermore, compared to HIRO, we do not generate sub-goals on inference; we do it only during training.

\section{Problem Statement}

We are interested in RL algorithms for navigation problems which consider safety constraints. To this end, we model our problem as a CMDP, where the agent must generate a sequence of actions that move the agent from the start state to the goal while avoiding static obstacles and ensuring a certain degree of safety.

\subsection{Navigation Problem} 

We are interested in robots whose dynamics are described by the control law $\dot{s} = f(s, a)$, where $s$ is the robot's state and $a$ is the robot's action. For the autonomous vehicle, for instance, we consider the bicycle model described in~\cite{surveyMotionPlanning}: $\dot{x} = v cos(\theta)\nonumber, \dot{y} = v sin(\theta), \dot{\theta} = \frac{v}{L} \tan(\gamma)$, where $x$,$y$ are the coordinates of the vehicle's rear axle, $\theta$ is the orientation, $L$ is the wheel-base, $v$ is the linear velocity, and $\gamma$ is the steering angle. 

All the robots operate in the 2D-3D workspace populated with static obstacles. 
Let $Obs$ denote the set of obstacles. Denote by $\mathcal{S}_{free}$ all the configurations of the robot that are not in collision with the obstacles. The problem is to find the sequence of actions $(a_1, \ldots, a_n)$ that move the agent from its start configuration $s_{start}$ to the goal $s_{goal}$, s.t. that every state from the resultant trajectory lies in $\mathcal{S}_{free}$. These actions are generated using a sequential decision-maker based on CMDP.

\subsection{Constrained Markov Decision Process}

Formally, CMDP can be represented as a tuple
$(\mathcal{S}, \mathcal{A}, \mathcal{P}, \mathcal{R}, \mathcal{C}_i,$ 
$d_i, \gamma)$, $i \in (1, \ldots, K)$ where $\mathcal{S}$ is the state space, $\mathcal{A}$ is the action space, $\mathcal{P}$ is the state-transition model, $\mathcal{R}$ is the reward function, $\mathcal{C}_i$ is a i-th constraint cost function and $\gamma$ is the discounting factor. Additionally, we consider a goal-conditioned setting of CMDP with the goal space $\mathcal{G} \subseteq \mathcal{S}$. During learning, at each time step the agent, being in state $s_t \in \mathcal{S}$ with a given goal $g_t \in \mathcal{G}$ takes action $a_t \in \mathcal{A}$ and receives a reward $r_t  = \mathcal{R}(s_t, a_t, s_{t+1}, g_t)$ and costs $c_{i} = \mathcal{C}_i(s_t, a_t, s_{t+1}, g_t)$.  The goal is to learn a policy, i.e., the mapping from the states to the distributions of actions, $\pi: \mathcal{S} \rightarrow P(\mathcal A)$. The policy should maximize the expected return $J(\pi)$ from the start state $s_t$ while satisfying the discounted cost with tolerable limit $d_i$ through the discounted costs $C_{i}(\pi)$ under policy $\pi$:
\begin{align*}
   &J(\pi) = \mathbb E_{\tau \sim \pi}[\sum_{i=0}^T \gamma^{t}r(s_t, a_t, s_{t+1}, g_t)],\\
   &C_{i}(\pi) = \mathbb E_{\tau \sim \pi}[\sum_{t=0}^T \gamma^{t}c_i(s_t, a_t, s_{t+1}, g_t)],
\end{align*}
where $\tau = (s_0, a_0, s_1, a_1, \ldots)$ denotes a trajectory.
The objective of CMDP for the policy $\pi$ is to find: 
\begin{equation}
    \pi^* = \arg\max_{\pi \in \Pi} J(\pi), \;\; s.t. \; C_i(\pi) \leq d_i, \forall i  
\end{equation}

We considered the current and goal states $s_t, g_t$ as a function of current and goal observations $s_t, g_t \approx  f^{\pi}(o_t, o^g_t)$ where $f^{\pi}$ are the lower layers of neural network approximator of the policy.

\section{Method}

We propose a novel learnable algorithm that comprises two reinforcement learning approaches (safe and hierarchical) to find a policy that solves long-horizon navigation tasks considering the limit of the cumulative safety constraints. On the one hand, our algorithm trains a subgoal policy that decomposes the initial problem into smaller sub-problems via intermediate goals (subgoals); on the other hand, it trains a safe policy that addresses the safety constraints. Overall, our algorithm generates subgoals while respecting the safety constraints, which helps to improve the safe policy exploration of our algorithm to solve the initially challenging task. We name this algorithm as SPEIS (Safe Policy Exploration Improvement via Subgoals). Now we will describe each policy separately and their integration (see Fig.~\ref{fig:method}).


\subsection{Safe Policy}

The safe policy is aimed at guaranteeing a degree of safety to avoid dangerous policy behavior. One of the most common approaches to address CMDP is the Lagrangian method in combination with the well-known off-policy gradient method SAC~\cite{haarnoja2018soft}. Our approach is based on the actor and critic architecture proposed in SAC and modified in SAC-Lagrangian~\cite{ha2021learning}.

\subsubsection{SAC} SAC is a value-based method that uses the maximum entropy principle. This principle encourages the agent to explore the environment to learn more about it. The SAC policy maximizes the expected return $J(\pi_{\theta})$ while the entropy constraint enforces a minimum degree of exploration $\mathbf{H}$.
\begin{equation}
    \begin{gathered}
    \max_{\theta} J(\pi_{\theta}), \; s.t. \; \mathbb{E}_{\tau \sim \pi_{\theta}} [ -\log(\pi_{\theta}(a_t|s_t, g_t))] \geq \mathbf{H}
    \end{gathered}
\end{equation}
where $\pi_{\theta}$ is an arbitrary differentiable policy. We refer our readers to \cite{haarnoja2018soft} for more details.

\subsubsection{SAC-Lagrangian} is a SAC-based method that, along with the maximum entropy term, also addresses the safety constraints through the Lagrangian method. 
\begin{equation}
    \begin{gathered}
    \max_{\theta} J(\pi_{\theta}), \; s.t. \; C_i(\pi_{\theta}) \leq d_i, \\
    \mathbb{E}_{\tau \sim \pi_{\theta}} [ -\log(\pi_{\theta}(a_t|s_t, g_t))] \geq \mathbf{H}
    \end{gathered}
\end{equation}

The Lagrangian methods allow the CMDP problem to be reduced to an unconstrained problem via Lagrange multipliers $gamma_i$ (safety weights), as shown in Eq.~\ref{eq:SACLagrangian_formula} via Lagrange multipliers $\lambda_i$ (safe weights). The latter is used to penalize the constraint violation between the tolerable limit $d_i$ and the discounted expected cost $C_i(\pi)$.
\begin{equation}\label{eq:SACLagrangian_formula}
   \min_{\lambda_i} \max_{\theta}L(\theta, \lambda_i) = J(\pi_{\theta}) - \sum_{i} \lambda_i (C_i(\pi_{\theta}) - d_i)
\end{equation}

Our approach: Unlike the approaches described above during the policy evaluation phase, we use two critics to estimate the action-value state for the reward and entropy term $Q^{\pi}_{r}$ and for the safety term $Q^{\pi}_{c}$ by minimizing the Bellman error with respect to the Q-function parameters $Q_{\phi}$ and $Q_{\psi}$ respectively. 
\begin{align}\label{eq:policy_evaluation}
\begin{split}
    Q^{\pi_{\theta}}_{r} = \arg \min_\phi \frac{1}{2} \mathbb{E}_{ (s_t, a_t, s_{t+1}, g_t) \sim D}[y^r_t -  Q^{\pi_{\theta}}_{r}(s_t, a_t, g_t)]^2 \\
    Q^{\pi_{\theta}}_{c} = \arg \min_{\psi} \frac{1}{2} \mathbb{E}_{ (s_t, a_t, s_{t+1}, g_t) \sim D}[y^c_t -  Q^{\pi_{\theta}}_{c}(s_t, a_t, g_t)]^2
\end{split}
\end{align}
with the target value for the reward and the entropy term and safety term given by:
\begin{align*}
    y^r_t = r(s_t, a_t, g_t) + \gamma \mathbb{E}_{a_{t+1} \sim \pi_{\theta}(.|s_t, g_t)} Q^{\pi_{\theta}}_{r} (s_{t+1}, a_{t+1}, g_{t+1}) \\
    y^c_t = c(s_t, a_t, g_t) + \gamma \mathbb{E}_{a_{t+1} \sim \pi_{\theta}(.|s_t, g_t)} Q^{\pi_{\theta}}_{c} (s_{t+1}, a_{t+1}, g_{t+1})
\end{align*}

\begin{figure*}[t]
    \centering
    \includegraphics[width=0.9\textwidth]{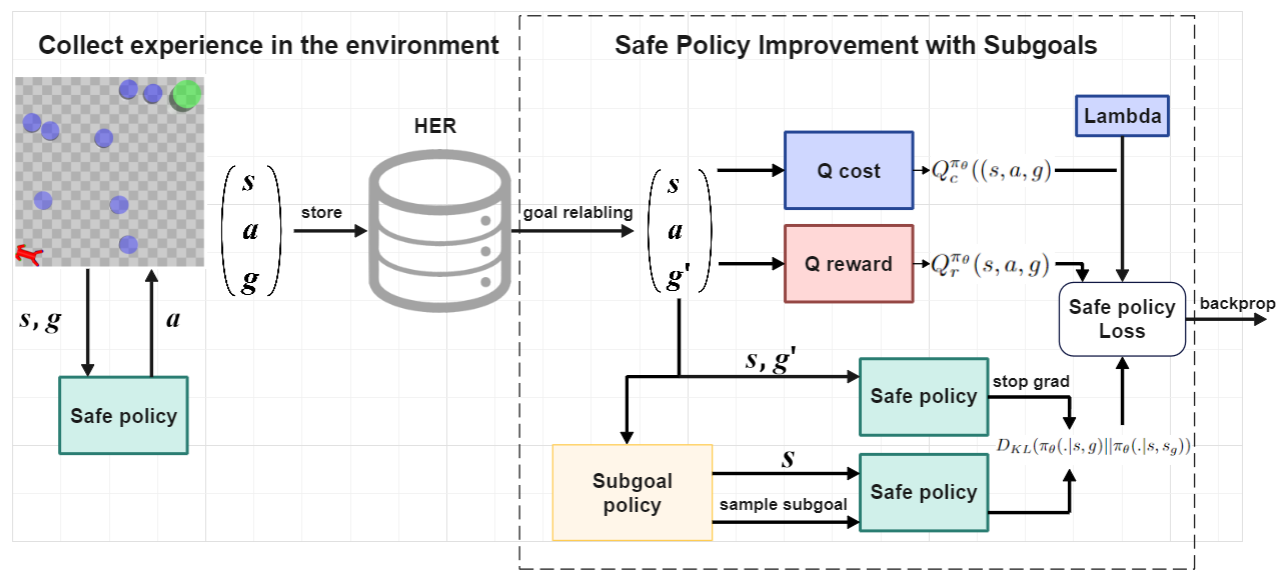}
    \caption{The transitions on safety-gym environment and policy update procedures.}
    \label{fig:method}
\end{figure*}

During the policy improvement phase, the actor $\pi_{\theta}$ is updated using the following actor loss function:
\begin{equation}
    \begin{gathered}
    J(\pi_{\theta}) = \mathbb{E}_{(s, g)\sim D}\mathbb{E}_{a\sim \pi_{\theta}(.|s, g)}[
    \beta \; \log(\pi_{\theta}(a|s, g)) - \\
    Q^{\pi_{\theta}}_{r}(s, a, g) + \lambda \; Q^{\pi_{\theta}}_{c}((s, a, g))]
    \end{gathered}
\end{equation}
where $D$ is the replay buffer. Simultaneously, we learn the adaptive Lagrangian multiplier (safe weight) $\lambda$ by minimizing the following loss:
\begin{equation}\label{eq:multiplier_optimization}
    \begin{gathered}
    J_s(\lambda) = \textrm{argmax }\mathbb{E}_{(s)\sim D}\mathbb{E}_{a\sim \pi_{\theta}(.|s)} [\lambda (d -  Q^{\pi_{\theta}}_{c}((s, a, g))],
    \end{gathered}
\end{equation} 
so the safe weight $\lambda$ will be decreased if $d \geq  Q^{\pi_{\theta}}_{c}$, otherwise $\lambda$ will be increased to further emphasize the safety.

\subsection{Hierarchical Policy}

The hierarchical method aims to facilitate learning of long-term horizon tasks. This hierarchical policy generates the intermediate goals (subgoals) between the start and goal states to simplify the original, challenging long-term horizon tasks. 

The approach with subgoals involves using two policies: hierarchical policy and main policy. The hierarchical policy predicts the subgoal while the main policy solves the main task: reaching the goal state from the start state. In general, various ways to implement the subgoals can be proposed. In this work, we suggest using the idea of the Imaged Subgoal policy proposed in RIS~\cite{chane2021goal} to predict subgoals in combination with a main policy (hereinafter referred to as safe policy). 

To predict the subgoals, the hierarchical policy minimizes the following costs:
\begin{equation}\label{eq:value_function}
    \psi_{\pi_{\theta}}(s_g|s, g) = \textrm{max}(|V^{\pi_{\theta}}(s, s_g)|, |V^{\pi_{\theta}}(s_g, g)|)
\end{equation}
where $s_g$ is the predicted subgoal and the norm of $V^{\pi_{\theta}}(s, s_g)$ is the expected discounted number of steps required for the main policy to reach $s_g$ from $s$.

\begin{algorithm}[t]
\caption{SPEIS}
\label{alg:POSCO}
\begin{algorithmic}[1]
    \State Initialize replay buffer D
    \State Initialize $Q_r, Q_c, \pi_{\theta}, \lambda, pi^{H}$
    \For{$k = 1, 2, \ldots$}
        \State Collect experience in D using $\pi_{\theta}$ in the environment
        \State Sample batch $(s_t, a_t, r_t, c_t, s_{t+1}, g) \sim D$ from HER
        \State Sample batch of subgoal candidates $s'_g \sim D$
        \State Update $Q_r, Q_c$ using Eq.~\ref{eq:policy_evaluation} (Policy Evaluation)
        \State Update $\lambda$ using Eq.~\ref{eq:multiplier_optimization} (Multiplier Optimization)
        \State Update $\pi^{H}$ using Eq.~\ref{eq:subogal_policy_improvement} (Subgoal Policy Improvement)
        \State Update $\pi_{\theta}$ using Eq.~\ref{eq:safe_policy_improvement} (Safe Policy Improvement with Subgoals see Fig.~\ref{fig:method})
    \EndFor
\end{algorithmic}
\end{algorithm}

The Kullback-Leibler divergence between subgoals and agent state distributions is used to generate valid imaged subgoals. Hence, the optimization task for hierarchical policy is:
\begin{equation}
    \begin{gathered}
    \pi^H = \textrm{argmax }\mathbb{E}\left[A^{\pi^H_{k}}(s_g|s, g)\right], \\
    \textrm{s.t.}\; D_{KL}(\pi^H(*|s, g)||p_s(*)) \leq \epsilon
    \end{gathered}
\end{equation}
where the advantage function is represented by:
\begin{equation}\label{eq:subogal_policy_improvement}
    A^{\pi^H_{k}}(s_g|s, g) = \mathbb{E}_{s'_g\sim \pi^H(*|s, g)}\left[\psi_{\pi_{\theta}}(s'_g|s, g) - \psi_{\pi_{\theta}}(s_g|s, g)\right]
\end{equation}

The subgoal policy $\pi^H$ uses only the main critic $V^{\pi}$ for updating and directs the training of the main policy using the subgoal generated for each transition. First, a transition $(s, a, r, s', g)$ is sampled from the replay buffer, and the hierarchical policy takes a sample from the subgoal $s_g\sim \pi^H(.|s, g)$. Secondly, the distribution $\pi_{\theta}(.|s, g)$ with respect to $g$ and the distribution $\pi_{\theta}(.|s, s_g)$ with respect to $s_g$ are calculated. Then the Kullback-Leibler divergence $D_{KL}(\pi_{\theta}(.|s, g)|| \pi_{\theta}(.|s, s_g))$ is used to shift the distribution given the goal $\pi_{\theta}(.|s, g)$ to the distribution given the subgoal $\pi_{\theta}(.|s, s_g)$. 
The policy optimization is provided by the following actor loss function:
\begin{equation}
    \begin{gathered}
    \pi_{\theta} = \textrm{argmax }\mathbb{E}_{(s, g)\sim D}\mathbb{E}_{a\sim \pi_{\theta}(.|s, g)} \\
    [Q^{\pi_{\theta}}(s, a, g) - \alpha D_{KL} (\pi_{\theta}(.|s, g)||\pi_{\theta}(.|s, s_g))]
    \end{gathered}
\end{equation}


\subsection{Safe Policy with Subgoals}

We propose an off-policy method SPEIS comprising hierarchical and safe policies, which alleviates the problem of the lack of safe exploration due to long-term horizon tasks and consideration of safety constraints. In this way, our algorithm acquires a safety behavior that reduces the cumulative safety constraints while maintaining a high reward thanks to the improvement of the safe policy exploration. Consequently, the policy can substantially decrease the collision rate while still providing a high success rate. 

Instead of using the principle of maximum entropy from SAC~\cite{haarnoja2018soft}, we use the hierarchical policy, which not only encourages the agent to explore the environment via subgoals but will also respect the safety constraints because the hierarchical policy is trained through the transitions that are generated by a safe policy. To this end, during the policy optimization, we add the divergence $D_{KL}$ to actor loss instead of the entropy term:
\begin{equation}\label{eq:safe_policy_improvement}
    \begin{gathered}
    J(\pi_{\theta}) = \mathbb{E}_{(s, g)\sim D}\mathbb{E}_{a\sim \pi_{\theta}(.|s, g)}Q^{\pi_{\theta}}_{r}(s, a, g) \\
    + \lambda Q^{\pi_{\theta}}_{c}((s, a, g)
    - \alpha D_{KL} (\pi_{\theta}(.|s, g)||\pi_{\theta}(.|s, s_g)))
    \end{gathered}
\end{equation}

The policy optimization is shown in Fig.~\ref{fig:method}. During this optimization, we try to find a balance between maximizing the reward $Q_r$ and reducing the violations of safety constraints $Q_c$ while guiding the policy through subgoals via $D_{KL}$ to achieve distant goals. Firstly, a transition $(s, a, r, c, s', g)$ is sampled from the replay buffer at each time step. Then, we calculate approximators $Q_r$ and $Q_c$ for the reward and safety terms using the Eq.~\ref{eq:policy_evaluation}. On the one hand, $Q_r$ encourages our policy to maximize the reward function, and $Q_c$ penalizes our policy for violating safety constraints. Simultaneously, we learn the adaptive Lagrangian multiplier $\lambda$ using the Eq.~\ref{eq:multiplier_optimization} as a safety weight for the $Q_c$. The safety weight is responsible for penalizing the violation of the cumulative safety constraints with respect to the limit $d_i$. Subsequently, the hierarchical policy takes a sample from the subgoal $s_g\sim \pi^H(.|s, g)$ to calculate the distribution $\pi_{\theta}(.|s, g)$ with respect to $g$ and the distribution $\pi_{\theta}(.|s, s_g)$ with respect to $s_g$ are calculated. Latter, the Kullback-Leibler divergence $D_{KL}(\pi_{\theta}(.|s, g)|| \pi_{\theta}(.|s, s_g))$ is used to shift the distribution given the goal $\pi_{\theta}(.|s, g)$ to the distribution given the subgoal $\pi_{\theta}(.|s, s_g)$. After that, we make the safe policy improvement using Eq.~\ref{eq:safe_policy_improvement}. As the implementation the safe policy is the probabilistic neural network parameterized by a squashed Gaussian distribution with diagonal variance $\pi_{\theta}(.|s, g) = tanh\;\mathcal{N}(\mu_{\theta}(s, g), \Sigma_{\theta}(s, g))$ and the subgoal policy is a probabilistic neural network parametrized by a Laplace distribution with diagonal variance $\pi(.|s, s_g)) = Laplace(\mu(s, g), \Sigma(s, g)$. The full algorithm is summarized in Algorithm~\ref{alg:POSCO}.

Unlike other algorithms~\cite{nachum2018data}, SPEIS uses a hierarchical policy which provides a subgoal $s_g$ with the same dimension of state $s$ containing the same information as the current observation, for example it may contain pseudo-lidar rays that can help the agent understand how close the agent is to obstacles. This allows our algorithm to be used not only in environments with a uni-task setting where the danger zone positions (or static obstacles) are fixed but also in a multi-task setting where the positions of hazard zones can be regenerated again in the new episode. Thanks to this, our policy can learn to reach more distant goals, while the reward value function becomes a better estimate for the distance between states and the cost value function becomes a better estimate to limit the exploration of dangerous states. This also allows the subgoal policy to propose more appropriate subgoals that guide the agent to reach more distant goals through safe exploration because the subgoal policy attempts to learn all the information from the observation.

\begin{table}[t]
\begin{center}
\scriptsize
\resizebox{0.99\linewidth}{!}{
\begin{tabular}{p{0.18\linewidth}|p{0.14\linewidth}|p{0.14\linewidth}p{0.14\linewidth}}
 Env & Robot & action dim & state dim \\
 \hline \hline
  \multirow{1}{4em}{POLAMP} & \multirow{1}{4em}{vehicle}
  & 2 & 88 \\
 \hline
  \multirow{1}{6em}{safety gym} & \multirow{1}{4em}{point}
  & 2 & 28 \\
 \hline 
  \multirow{1}{6em}{safety gym} &\multirow{1}{4em}{car}
  & 2 & 42 \\
\hline 
  \multirow{1}{6em}{safety gym} &\multirow{1}{4em}{doggo}
  & 12 & 108 \\
\hline 
  \multirow{1}{6em}{safety gym} &\multirow{1}{4em}{sweeps}
  & 2 & 4 \\
 \hline
 \hline
\end{tabular}
}
\caption{The dimensions of each robot's action space and state space are provided for POLAMP and safety-gym environments.}
\label{tab:StateActionDim}
\end{center}
\end{table}

\section{Experiment Setup and Configuration}\label{Ch:experiment_setup}

\subsection{Environments}
We considered two Gym environments to train and evaluate our method: Safety-Gym and POLAMP.

\begin{figure}[t]
    \centering
        \includegraphics[width=0.1\textwidth]{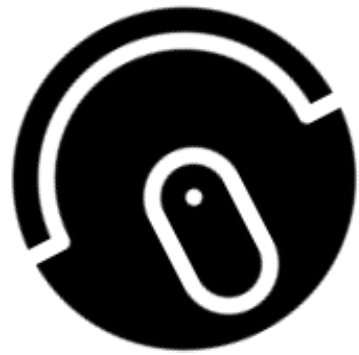}
        \includegraphics[width=0.1\textwidth]{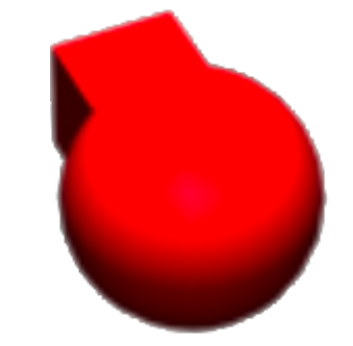}
        \includegraphics[width=0.1\textwidth]{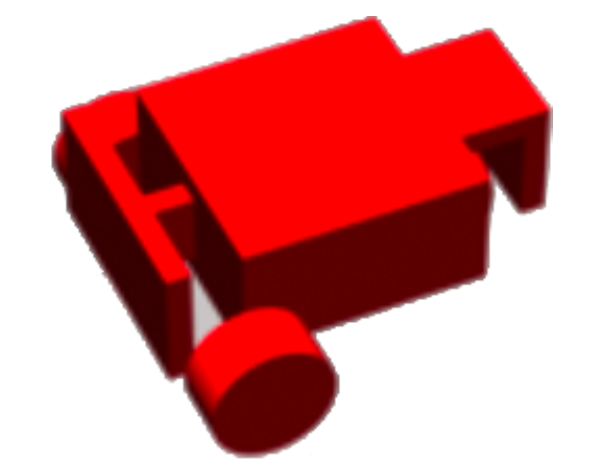}
        \includegraphics[width=0.1\textwidth]{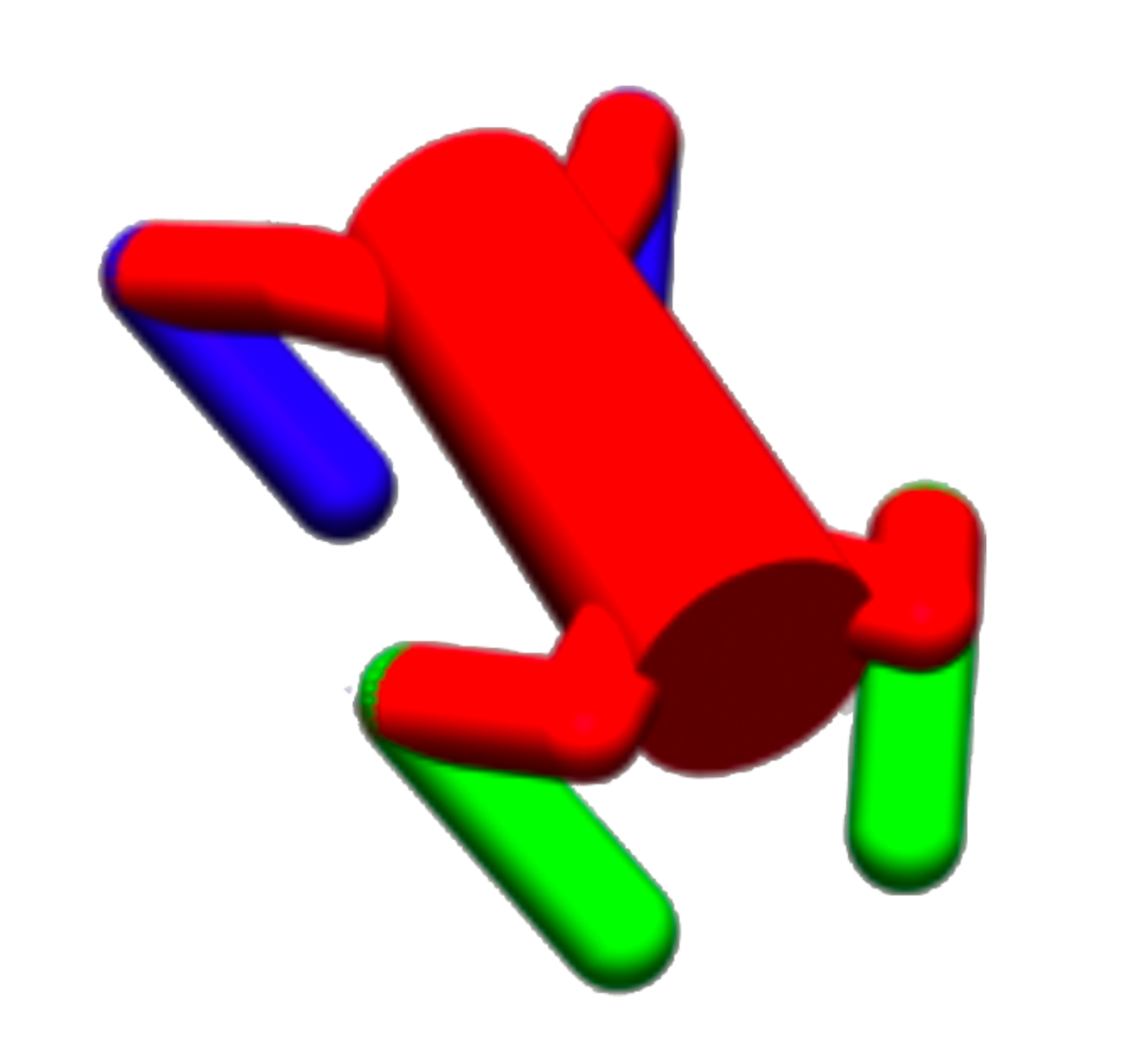}
    \caption{Robots from safety-gym environment: sweeping, point, car, doggo.}
    \label{fig:robots}
\end{figure}

\paragraph{Safety-gym environment}
The safety-gym environment is proposed in~\cite{ray2019benchmarking} and is a well-known environment among the RL community that for evaluating RL methods that respect safety constraints. This Gym environment provides different types of robots that are shown in Fig.~\ref{fig:robots}: Point, Doggo, and Car are from~\cite{ray2019benchmarking} and Robot Sweep from~\cite{roza2022safe}. Point models a robot with engines to rotate and move forward, Car has two independently operated parallel wheels with a rear-wheel axle for balance, Quadruped models a quadrupedal robot with each leg having torque controls in the hip and knee joints and Sweep directly controls its x, y coordinates so it is a holonomic robot. All agents obtain their state information from the joints, accelerometer, gyroscope, magnetometer, velocimeter, and a 2D vector pointing toward the goal. The robot is surrounded by static obstacles(hazard zones), and we partially utilize the information around the robot through the closest 4 obstacles to the robot. Considering a goal-conditioned configuration, the observation consists of the vector based on the current robot's state and the goal one. The dimensions of each robot's action and observation space are presented in Tab.~\ref{tab:StateActionDim}. For details of the actions and kinematic model, we refer the readers to ~\cite{ray2019benchmarking}.

\begin{figure}[ht!]
    \centering
    \includegraphics[width=0.21\textwidth]{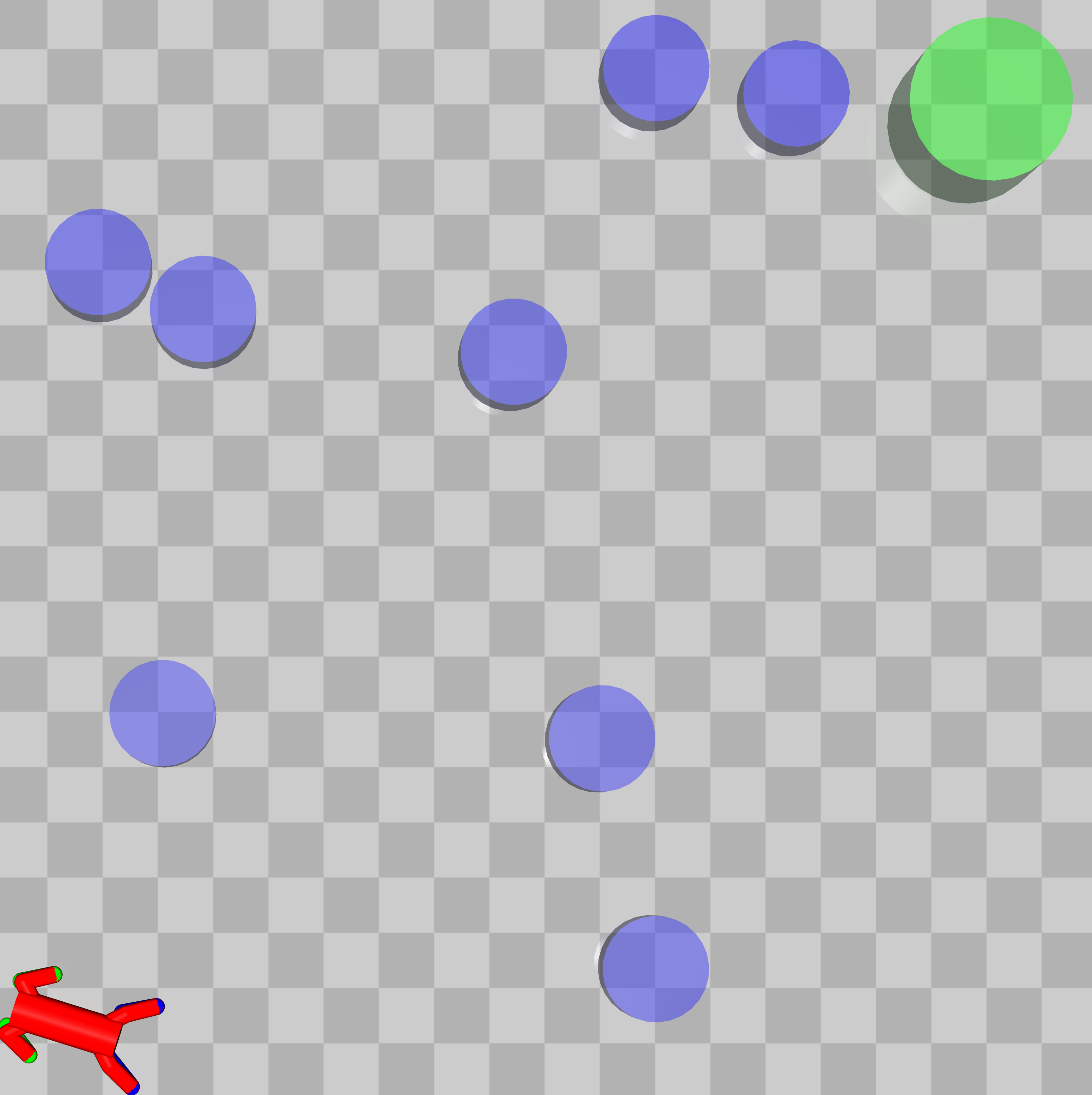}
    \includegraphics[width=0.25\textwidth]{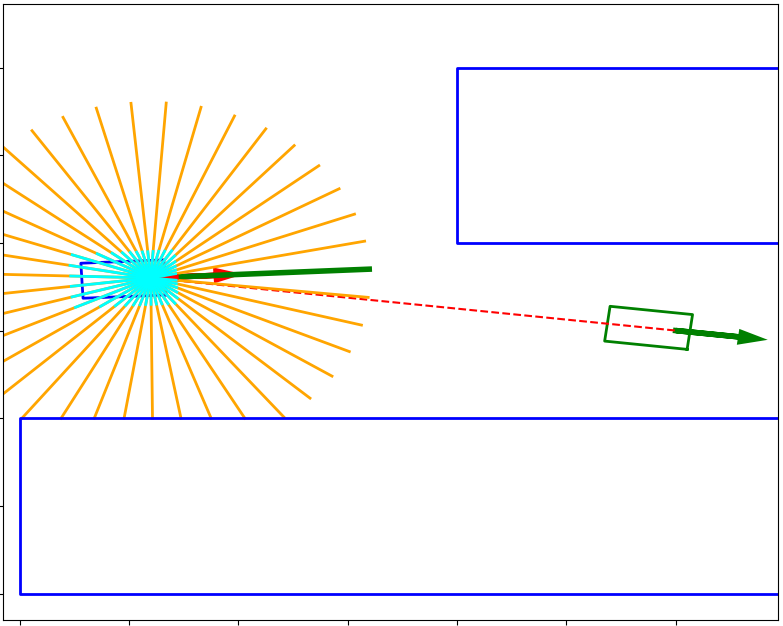}
    \caption{On the left is shown the safety-gym environment, where the blue cylinders represent hazardous zones, the green cylinder -- the goal and the quadruped robot is represented by red. On the right -- POLAMP environment, where the blue rectangles represent the static obstacles, the orange arrays -- pseudo lidars, the cyan arrays -- safe zone, the green rectangle with an arrow -- the goal one, while the blue rectangle with a red arrow represents the vehicle}
    \label{fig:environments}
\end{figure}

\paragraph{POLAMP environment} 
The POLAMP environment is proposed in~\cite{angulo2022policy} and provides a simulator with an autonomous vehicle equipped with a kinematic bicycle model and lidar sensor. This simulator is populated with static obstacles(hazard zones) as shown in Fig.\ref{fig:environments}.

This paper considers the same action space as~\cite{angulo2022policy}. The actions $a_t = (a, \omega)$ are composed of the linear acceleration $a$ and rotation rate $\omega$. We refer the readers to ~\cite{angulo2022policy} for details of the actions and kinematic model. The observation $o_t$ is a vector that consists of the $N_{beams}=39$ measurements of the lidar that cover the 360$^{\circ}$ surrounding of the robot up to the length of $beam_{max} = 10\;m$ concatenated with the robot's state  $(x, y, \theta, v, \gamma)$, where $(x, y, \theta, v, \gamma)$ are the parameters of the current state. We consider an ideal environment, so both the simulation and actuation models.

\subsection{Policy Training}
we describe the details of policy training for the considered environments.
\paragraph{rewards and safety constraints}\label{paragraph:observations-actions-reward}


The sparse reward function is described by $\mathcal R = w_r^T [r_{\text{t}}, r_{\text{col}}]$ where $w_r$ is a vector of weights, $r_{\text{t}}=-1$ is the constant penalty for each time step and $r_{\text{col}}$ is $-1$ if the agent collides with an obstacle and 0 otherwise. We set the weights to be $w_r = (1, 100)$. Unlike the other method~\cite{chane2021goal, nachum2018data} where the agent could collide with obstacles and continues trying to reach the goal receiving $r_{\text{t}}$ as a penalize, in this work we immediately end the episode when the agent collides with obstacles to provide safer behavior penalizing the agent with $r_{\text{col}}$. We consider that the value of $V^{\pi_{\theta}}(s, s_g)$ in the Eq.~\ref{eq:value_function} should be bigger than usual, suggesting that the robot ends the episode far from the goal when the agent collides with obstacles. On the other hand, when the robot does not collide the $V^{\pi_{\theta}}(s, s_g)$ retains the same meaning as the expected discounted number of steps required for the main policy to reach the subgoal $s_g$ from the state $s$.

The safety constraint cost is induced by a boolean value that evaluates the safety of the state at each timestamp by indicating whether or not the autonomous vehicle moves near to the obstacles, as proposed in \cite{chow2019lyapunov}. We impose an immediate constraint cost as: $c_i(s_i, a_i, g_i) = 1\{\boldsymbol{l}_{min} \leq r_{safety}\}$ where $r_{safety} = 0.3$m is the safety radius, $\boldsymbol{l}_{min}$ is the minimum current lidar signal (or the minimum robots's distance to obstacles for environments without lidars), and $1$ is the boolean value.

This safety constraint allows to the agent to move away from the obstacles so as not to violate the tolerable limit $d_i$.

\paragraph{HER Replay Buffer\label{policyLearning}}

For the policy training we use the HER replay buffer \cite{andrychowicz2017hindsight} which allows the agents to cope with the sparse rewards. HER uses the fact that even if a desired goal was not achieved, another goal may have been achieved during a rollout. The latter could help the agent to reach different goals through virtual transitions, i.e. by relabeling the initial goal with another goal that is a sample of the transitions found in the replay buffer. The goal relabeling strategy from the original article \cite{andrychowicz2017hindsight} was used:
\begin{itemize}
    \item 20\%: original goals from collected trajectories,
     \item 40\%: randomly sampled states from the replay buffer trajectories,
      \item 40\%: future states along the same collected trajectory.
\end{itemize}

\paragraph{Dataset}

\begin{figure}[t]
  \begin{center}
\includegraphics[width=0.8\linewidth]{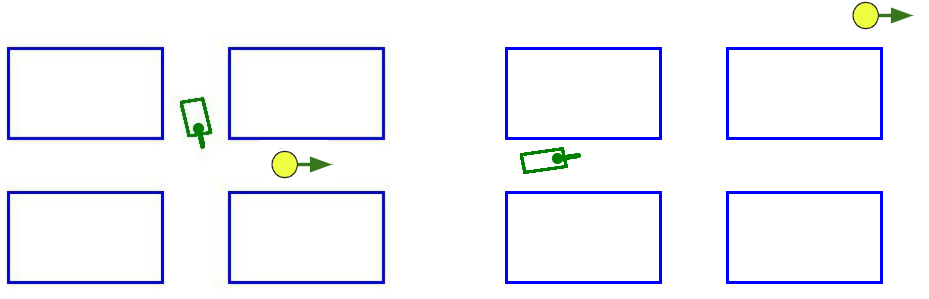}
  \end{center}
\caption{Example dataset patterns. The left figure illustrates one of patterns from the level 1 and the right figure illustrates one of patterns from the level 2, which were used for training and validation.}
\label{fig:learning_dataset}
\end{figure}

For POLAMP, we created two dataset with different level of complexity in static environment with challenging narrow corridors that has a size $40m \times 40m$. The level 2 consists of challenging tasks where the agent should turn twice in narrow corridors to reach the goal and the level 1 has simpler tasks where the agent should at most turn once. The level 1 and level 2 consist of different tasks (start and goal states) as illustrated by the examples shown in Fig.~\ref{fig:learning_dataset}. To generate tasks for level 2 in static environments we choose 12 patterns to generate the start and goal states randomly such that the distance between the start and goal locations was in the interval of $[20, 30]$m. For level 1 we choose 30 patterns to generate the start and goal states randomly such that the distance between the start and goal locations was in the interval of $[10, 20]$m.

For the safety gym environment we use the default random generation of all hazard obstacles and tasks. The environment has a size $4m \times 4m$ and constantly has 8 hazard obstacles with circular shape. Unlike POLAMP environment in safety gym environments, every new episode, static obstacles (hazard zones) are randomly generated again. After that, the task (start and goal state) is generated so that the inflated shape of the robots in these states does not collide with obstacles.

\section{Experimental Evaluation}

We evaluated our method and compared it with the competitors in a popular safety gym environment (see Fig.~\ref{fig:method}) and a complex environment with narrow corridors (see Fig.~\ref{fig:learning_dataset}).

\begin{figure*}[t]
    \centering
        \includegraphics[width=0.95\textwidth]{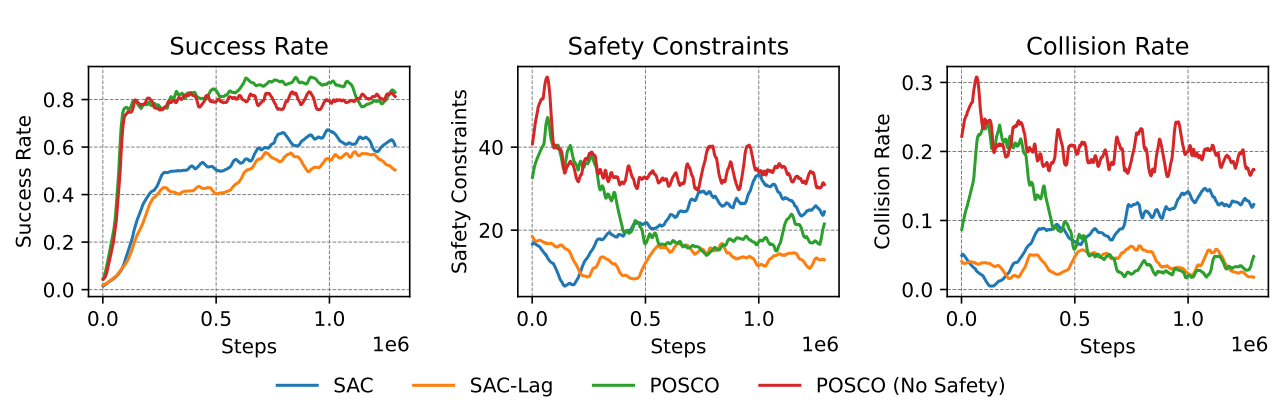}
    \caption{The comparison of policy training between SAC, SAC-LAG, SPEIS-noSAFETY and SPEIS for point environment.}
    \label{fig:ablation_study}
\end{figure*}

\paragraph{Ablation Study}
We conducted several experiments showing how the safety and hierarchical methods influence policy performance. We compared SPEIS to SAC, SAC-Lagrangian, and SPEIS-noSAFETY (see Fig.~\ref{fig:ablation_study}). We denoted the original RIS method as \textbf{SPEIS-noSAFETY (without safety)} to more clearly show the influence of the safety policy. SAC-Lagrangian and SPEIS provide a certain degree of safety by penalizing violations of safety constraints during training. SPEIS-noSAFETY and SPEIS use a subgoal policy to deal with the long-term horizon problem. We use a fixed entropy coefficient of $H=0.2$ for the SAC policy. As a limit of safety constraints, we set $d_i = 3.0$ with the Lagrangian initialization $\lambda_0 = 0.5$. For the hierarchical part, we establish the influence of the subgoal $\alpha = 0.5$. The task was considered solved if the agent reached the goal state with the Euclidean error $\epsilon_{\rho} \leq 0.3$ m with no collisions.

\begin{figure}[t]
  \begin{center}
\includegraphics[width=0.8\linewidth]{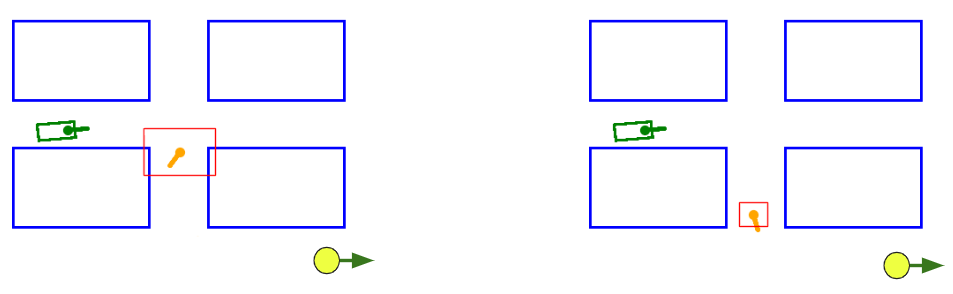}
  \end{center}
\caption{Predicted subgoals of SPEIS-noSAFETY \textbf{in the left figure} and SPEIS \textbf{in the right figure}. \textbf{Orange arrow with the rectangles} represent the subgoals, where the arrows represent the mean of the coordinates (x, y), and the rectangles cover the standard deviation of these coordinates. The standard deviation of one SPEIS subgoal is smaller than the other, i.e., the subgoal becomes safer due to the safety constraints.}
\label{fig:comparison_of_subgoals}
\end{figure}

The first clear trend that we can see is that the consideration of safety constraints provides the policy a safe behavior, as it tends to reduce the sum of safety constraints, as seen in Fig.~\ref{fig:ablation_study}b, which in turn leads to a substantial reduction in collisions, as shown in Fig.~\ref{fig:ablation_study}c. On the other hand, methods that use a subgoal policy increase the success rate by up to 40\%, as shown in Fig.~\ref{fig:ablation_study}a. One of the most interesting effects of our proposed method is that not only does the safe policy acquire safer behavior, but even the subgoal policy as it generates safer subgoals (see Fig.~\ref{fig:comparison_of_subgoals}). The standard deviation of one SPEIS subgoal is smaller than the SPEIS-noSAFETY, i.e., the subgoal becomes safer due to the safety constraints because the subgoal policy generates samples from the distribution $s_g = Laplace(\mu(s, g), \Sigma(s, g)$ that are more distant from the obstacles.

\paragraph{Evaluation in Safety Gym Environments}
A separate set of validation tasks was created, in a manner similar to the training dataset, and used to measure the progress of training, i.e., once in a while, we evaluated the performance of the currently trained policy on the validation tasks. After training, we evaluate every policy on a validation dataset through the following metrics: success rate (SR, \%), collision rate (CR, \%), and meantime safe cost (MTSC, t) of the successfully generated trajectories, i.e., how many time steps is the robot near to the obstacles, see safety constraints ~\ref{Ch:experiment_setup}.a. 

We compared SPEIS to the following algorithms: SAC, SAC-Lagrangian, SPEIS-noSAFETY, Lyapunov-RRT~\cite{xiong2022model} and CPO~\cite{achiam2017constrained}. All baselines are end-to-end RL policies except Lyapunov-RRT, which combines RRT* planner \cite{karaman2011anytime} with Lyapunov Neural policy. Unlike the other methods, it assumes full observability. We used the same learning setup for all the learning algorithms except for CPO, which does not use sparse reward. For CPO, we use the reward function from \cite{ray2019benchmarking}.

Colearning of the policy and Lyapunov function was conducted in an obstacle-free environment with state observation $(x, y, \theta, v, \delta)$. For testing, we utilized the geometric RRT* algorithm from article~\cite{xiong2022model} with the following parameters: maximum number of iterations $N_{max}=18000$, search step size $step_{search}=0.1$, and maximum step size $step_{max}=5$. For each of the 96 tasks at levels 1 and 2, the experiments were conducted 5 times to take into account the randomness of the RRT* algorithm.

The results are presented in Tab.~\ref{tab:resultsForSafetyGymEnvironment}. The first clear trend is that SPEIS has the lowest collision rate within the end-to-end RL algorithms while having one of the highest success rates. Only SAC-Lagrangian has a slightly lower collision rate, but that is because the SAC-Lagrangian agent prefers to stay in place instead of trying to reach the goal. On the contrary, the SPEIS agent always tries to reach the goal, and only when it enters dangerous zones will it prefer to stay in place at a safe distance from the obstacles instead of colliding with an obstacle. 

Our method provides a lower cumulative safety cost than Lyapunov-RRT because it was trained not to violate its limit during the training. Lyapunov-RRT provides a higher success rate since it utilizes all the information about the environment using RRT* to generate a collision-free path, while our method SPEIS is trained with a partial observation. Lyapunov-RRT tends to generate paths that avoid accumulating obstacles to provide a more comfortable initial path that follows the policy. However, this effect increases the number of steps needed to reach the goal, and this behavior is not effective. Also, Lyapunov-RRT provides higher cumulative safety constraints since the Lyapunov policy is unaware of obstacles and can approach them and even collide with them. On the other hand, the SPEIS agent is aware of obstacles and can stay in place if it is considered dangerous to move forward to prevent collisions. Furthermore, our algorithm has a lower time to inference with respect to Lyapunov-RRT because the latter generates a Lyapunov table and creates a collision-free plan based on RRT* for each task in addition to action prediction and modeling the robot's movement.

\begin{figure}[t]
    \centering
    \includegraphics[width=0.23\textwidth]{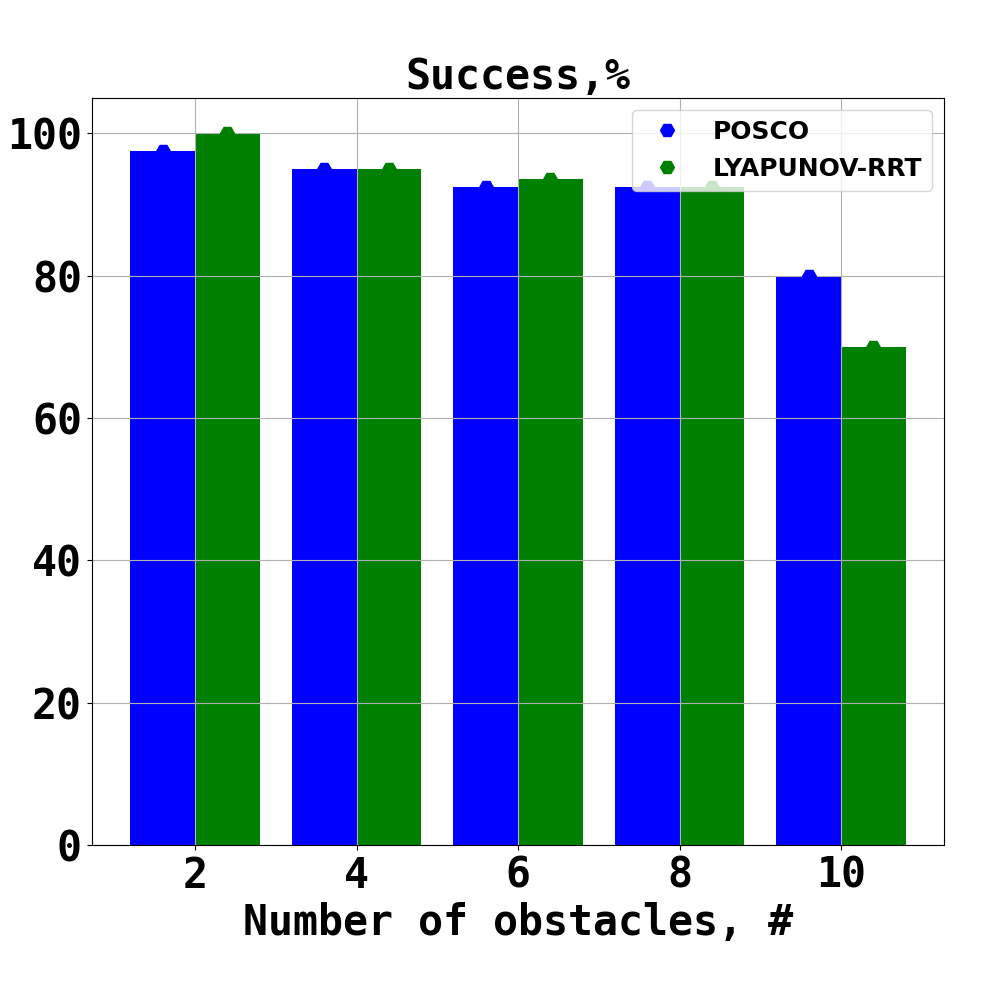}
    \includegraphics[width=0.23\textwidth]{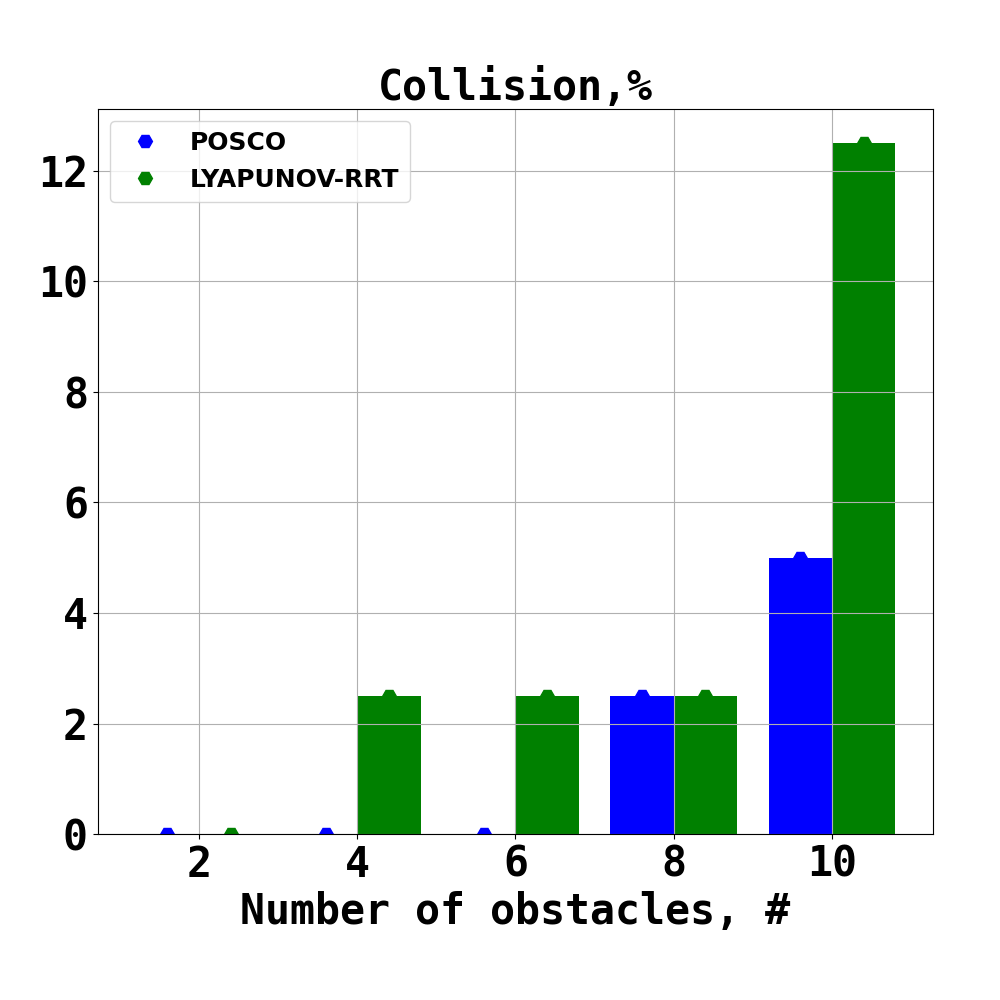}
    \caption{A comparison between SPEIS and Lyapunov RRT with different numbers of dynamic obstacles.}
    \label{fig:comparison_of_posco_and_lyapunov_rrt}
\end{figure}

To understand the generalization of our method and Lyapunov-RRT, we conducted an additional experiment (see Fig.~\ref{fig:comparison_of_posco_and_lyapunov_rrt}) with different numbers of static obstacles $N_{obs} = {2, 4 \ldots, 10}$. The first trend we can see is that our method provides fewer collisions while none are observed for up to 6 obstacles. On the other hand, Lyapunov-RRT provides more collisions than SPEIS, generating collisions with as few as 4 obstacles. One of the reasons is that the Lyapunov agent is not aware of the obstacles and finds it more difficult to follow the geometric path, avoiding collision with obstacles, while our agent is aware of the obstacles and tries not to collide thanks to the awareness of safety constraints. Although Lyapunov-RRT provides a slightly higher success rate for up to 6 obstacles, with 8 and 10 obstacles, we provide a higher success rate. The main reason is that Lyapunov provides less safe zones through which the agent can move, and thus, it is more difficult to reach the desired goal. The other reason is that it is more difficult to find a suitable geometric path for Lyapunov's policy when the number of obstacles increases. This trend shows that the performance of the Lyapunov-RRT highly depends on the complexity of the environment.

Overall, the performed experiments show that our policy SPEIS provides safer behavior compared to all the baselines (including the upper-bound Lyapunov-RRT), reducing the collision rate to a minimum while providing a higher success rate among RL end-to-end methods. Although it provides a slightly lower success rate than Lyapunov-RRT, the inference time is better, and the episode duration is shorter.

\paragraph{Evaluation in Narrow Environment}
We compared SPEIS to the following end-to-end RL algorithms: SAC, SAC-Lagrangian, and SAC Neural Conditioned (SAC-NCMP)~\cite{huang2021risk}. The latter is based on the Chance-Constrained Markov Decision that uses risk-bounded probabilistic constraints. It means that at each time step, we sample $N$ action noises from a normal distribution $\mathcal{N}(0, 1)$ to add to the current action, and we calculate how many of them violate the safety constraints $N_{v} / N$. We trained the policy with upper risk bound $\Delta = 0.3$, i.e., we consider that the state is not safe if the probability of violation of the constraints is more than $\Delta$. We also compared SPEIS to the hybrid algorithm Lyapunov-RRT and the classical algorithm RRT with Dubins steering function~\cite{DUBINS}.

The results between end-to-end RL algorithms are presented in Tab.~\ref{tab:resultsForPolampEnvironment}. The first clear trend is that SPEIS in level 2 complexity has a higher success rate while it has a lower cumulative safety cost. The higher success rate of SPEIS is because it uses a hierarchical policy that allows it to solve the most challenging tasks. The lower cumulative safety cost is due to the fact that the policy reached convergence while others did not. 

\begin{table}[t]
\begin{center}
\resizebox{0.99\linewidth}{!}{
\begin{tabular}{p{0.09\linewidth}|p{0.28\linewidth}p{0.14\linewidth}p{0.14\linewidth}p{0.14\linewidth}p{0.14\linewidth}}
 Env & Policy & SR,\% & CR,\% & MTSC, t\\
 \hline \hline
  \multirow{2}{4em}{point}
  & SPEIS & 92.5 & 2.5 & 20.7\\
  & SPEIS-noSAFETY & 88.6 & 11.1 & 33.2\\
  & L-RRT & \textbf{100} & 0 & 42.2 \\
  & SAC-LAG & 53.6 & 3.5 & \textbf{13.4} \\
  & SAC & 61.3 & 12.3 & 26.3 \\
  & CPO & 60 & 27.5 & 146.1 \\
 \hline 
  \multirow{2}{4em}{car}
  & SPEIS & 74.8 & 5.1 & 20.7\\
  & SPEIS-noSAFETY & 75.8 & 23.9 & 33.4 \\
  & L-RRT & \textbf{95.0} & 5.0 & 37.6 \\
  & SAC-LAG & 56.2 & 4.1 & \textbf{13.8} \\
  & SAC & 30.1 & 77.6 & 17.3 \\
  & CPO & 62 & 32.5 & 133.6 \\
\hline 
  \multirow{2}{4em}{doggo}
  & SPEIS & 79.2 & 5.8 & \textbf{12.5}\\
  & SPEIS-noSAFETY & 74.1 & 20 & 40.2\\
  & L-RRT & \textbf{92.5} & 7.5 & 17.1 \\
  & SAC-LAG & 2 & 10 & 22.6 \\
  & SAC & 8 & 15 & 44.63 \\
  & CPO & 65 & 30 & 86.8 \\
\hline 
  \multirow{2}{4em}{sweeps}
  & SPEIS & \textbf{100} & 0 & \textbf{0.98} \\
  & SPEIS-noSAFETY & \textbf{100} & 0 & 12.87 \\
  & L-RRT & \textbf{100} & 0 & 9.0 \\
  & SAC-LAG & 47 & 0 & 1.25 \\
  & SAC & 55 & 5 & 5.16 \\
  & CPO & 15 & 85 & 102 \\
 \hline
 \hline
\end{tabular}
}
\caption{Results of the experiments for Safety Gym environments.}
\label{tab:resultsForSafetyGymEnvironment}
\end{center}
\end{table}

The poor performance of SAC-NCMP in comparison with SPEIS is because even a little noise in narrow corridors causes the agent to violate the upper probabilistic risk bound. The agent believes that he violates safety constraints when he does not, making the training difficult. A big drawback of this approach is that the training takes much longer than other policies because of the number of samples in each timestep.

SAC-LAG shows a good performance on the dataset of level 1, but it has zero performance on the dataset of level 2. The lack of exploration to address the safety constraints and the long-term horizon task causes the policy to fail to solve the most challenging tasks. As we noticed during the training, the SPEIS converges faster in comparison with the SAC-LAG. Thanks to the subgoal policy, our method learns faster and solves more complex tasks, while the SAC-LAG without the subgoal policy cannot do it. While SPEIS respects the available limit of cumulative safety constraints, its performance is similar to SPEIS-noSAFETY, which does not consider them. 

 \begin{table}[t]
\begin{center}
\resizebox{0.99\linewidth}{!}{
\begin{tabular}{p{0.06\linewidth}|p{0.25\linewidth}p{0.14\linewidth}p{0.14\linewidth}p{0.14\linewidth}p{0.14\linewidth}}
 
 Level & Policy & SR,\% & CR,\% & MTSC, t\\
 \hline \hline
  \multirow{2}{4em}{1}
  & SPEIS & \textbf{92} & \textbf{1} & \textbf{1.14}\\
  & SAC-NCMP & 87 & 2 & 6.52 \\
  & SAC-LAG & 81 & 4.1 & 2.86 \\
  & SAC & 80.2 & 4.1 & 2.90 \\
 \hline 
  \multirow{2}{4em}{2}
  & SPEIS & \textbf{84} & \textbf{2} & \textbf{3.97}\\
  & SAC-NCMP & 33 & 21 & 66.58\\
  & SAC-LAG & 0 & 0 & 0 \\
  & SAC & 0 & 0 & 0 \\
 \hline
 \hline
\end{tabular}
}
\caption{Results of the experiments for POLAMP environments.}
\label{tab:resultsForPolampEnvironment}
\end{center}
\end{table}

The results of the comparison with algorithms based on the RRT algorithm are presented in Tab.~\ref{tab:comparison_with_rrt_based_algorithms}. Additionally, we evaluate these algorithms through the following metrics: time-to-reach (TTR, \%) and Planning time (PT, s). Time-to-reach is the average time steps it takes the agent to reach the goal, and planning time is the time it takes the RRT algorithm to generate the route.

We compared SPEIS to RRT with Dubins car~\cite{DUBINS} and Lyapunov-RRT. RRT algorithms assume full observability, unlike our method. The first clear trend is that RRT with Dubins car has a higher success rate. However, this algorithm does not take into account safety constraints and generates trajectories very close to obstacles. Besides, the Dubins car is able to generate only kinematic actions $v, \gamma$, while our method generates kinodynamic actions $a, \omega$ which are crucial for following the trajectory.

The high collision rate of the LyapunovRRT algorithm on dataset level 2 is explained by the default geometric RRT* not generating a kinematically feasible path. Given the narrow corridors in the environment POLAMP, the radius ROA used in the LyapunovRRT algorithm becomes very small, requiring the agent to move along the planned path with a small error. Consequently, due to the kinematic infeasibility of such a path, the agent cannot follow the plan and solve the task in most cases.

Unlike RRT-based algorithms, our end-to-end RL algorithm SPEIS does not require trajectory generation, which is typically time-consuming and can provide a high success rate while providing lower cumulative safety cost, as shown in Table ~\ref{tab:comparison_with_rrt_based_algorithms}. Furthermore, unlike RRT algorithms, SPEIS reaches the goal faster because it generates optimal actions to reach it, while RRT algorithms rely on randomness and generate suboptimal trajectories as shown by the TTR metric in Table ~\ref{tab:comparison_with_rrt_based_algorithms}.

Overall, the experiments that were conducted show that our policy SPEIS solves the navigation tasks well in level 2, respecting the safety constraints. Combining hierarchical and safety policies shows potential for working in challenging environments with narrow corridors.

\begin{table}[t]
\begin{center}
\resizebox{0.99\linewidth}{!}{
\begin{tabular}{p{0.06\linewidth}|p{0.1\linewidth}p{0.05\linewidth}p{0.05\linewidth}p{0.12\linewidth}p{0.12\linewidth}p{0.12\linewidth}p{0.12\linewidth}
}
 
 Level & Policy & SR,\% & CR,\% & MTSC, t & TTR, t & PT, s 
 \\
 \hline \hline
  \multirow{2}{4em}{1}
  & SPEIS & 
  92 & 
  \textbf{1} & \textbf{1.14} & \textbf{47.52} & \textbf{0} 
  \\
  & L-RRT 
  & 60 & 
  24 & 
  1.3 & 
  104.2 & 
  10.4 
  \\
  & RRT & 
  \textbf{100} & 
  0 & 
  2.1 & 
  86.3 & 
  18.3 
  \\
 \hline 
  \multirow{2}{4em}{2}
  & SPEIS & 
  84 & 
  \textbf{2} & \textbf{3.97} &
  164.53125 & \textbf{0} 
  \\
  & L-RRT & 
  16 & 
  78 & 
  15.1 & 
  170.1 & 
  30.2 
  \\
  & RRT & 
  \textbf{100} & 
  0 & 
  9 & 
  \textbf{111.4} & 
  51.4
  \\
 \hline
 \hline
\end{tabular}
}
\caption{Results of the experiments for POLAMP environments with classic planners.}
\label{tab:comparison_with_rrt_based_algorithms}
\end{center}
\end{table}

\section{Conclusion}

In this paper, we examined the navigation problems that involve distant goals and the requirement to adhere to safety constraints. To address this type of problem, we proposed a method comprising safe and hierarchical algorithms that jointly provide an end-to-end learnable policy -- SPEIS. Our method was empirically evaluated with 5 types of robots in two different environments, Safety-Gym and POLAMP, demonstrating its superiority over all state-of-the-art end-to-end RL approaches in terms of cumulative safety constraints. We also showed approximately the same results in environments of varying complexity (challenging environments with narrow corridors and safety-gym environments), while non-learnable competitors show worse performance in challenging environments with narrow corridors.

One of the future directions is to introduce a model-based algorithm into our method to provide safer behavior and more efficient safe exploration through the environment and to show better performance.

\bibliographystyle{root}
\bibliography{root}




\newpage


\begin{IEEEbiography}[{\includegraphics[height=1.1in,clip,keepaspectratio]{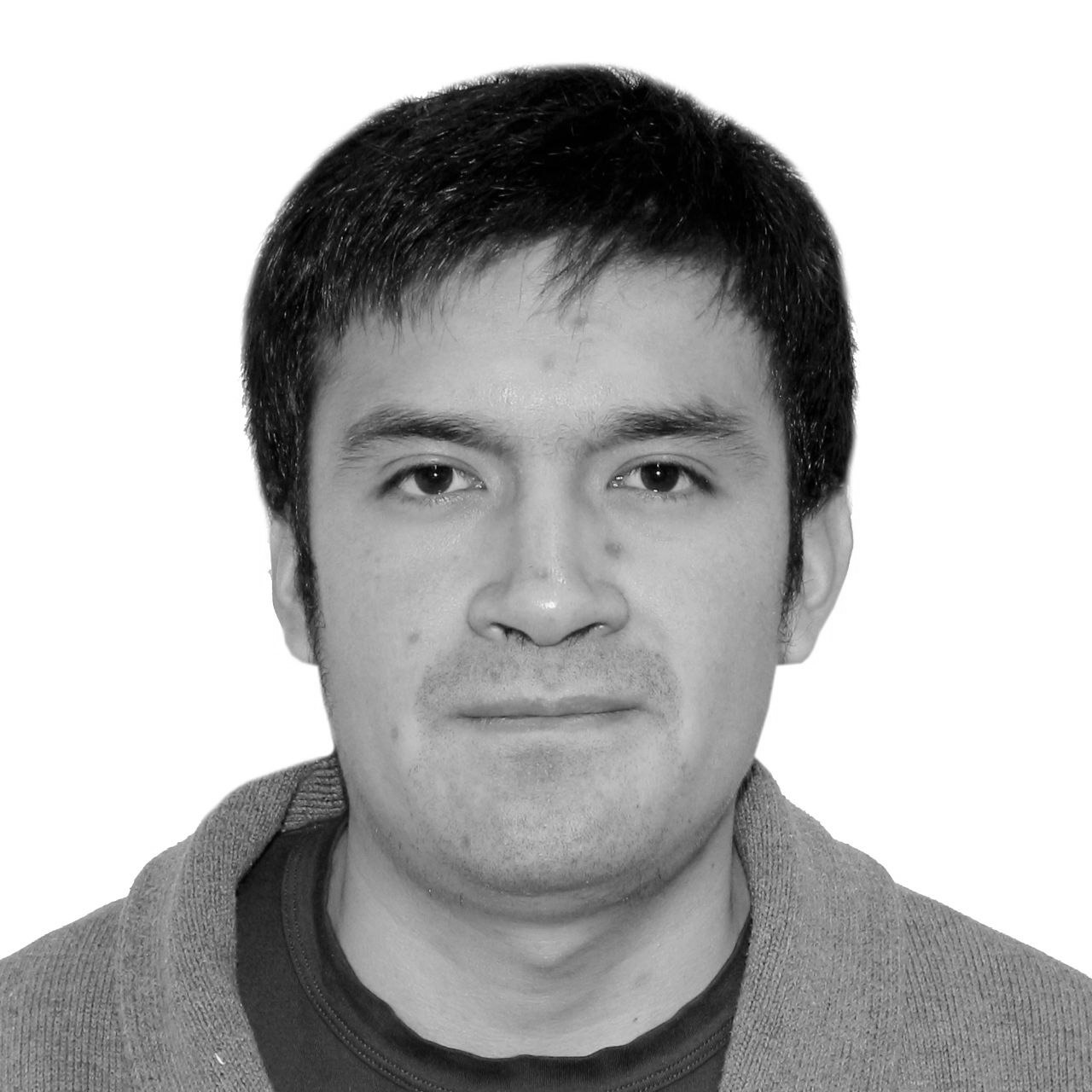}}]{Brian Angulo}
received a Master's degree from the Moscow Institute of Physics and Technology, Moscow, Russia, in 2020.

He is currently a PhD student at the Moscow Institute of Physics and Technology and a Leading Researcher at the robotics company OOO ``IntegraNT``. His research interests include reinforcement learning, motion planning, and robotics.

\end{IEEEbiography}

\begin{IEEEbiography}
[{\includegraphics[width=1in,height=1.5in,clip,keepaspectratio]{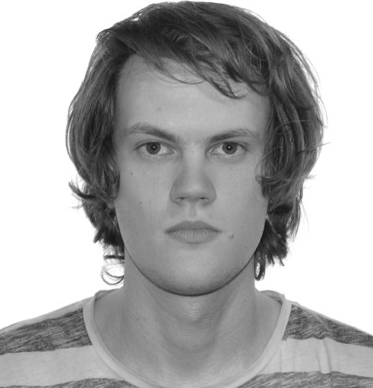}}]{Gregory Gorbov} obtained a Master's degree from the Moscow Institute of Physics and Technology in 2023 in the program of Methods and Technologies of Artificial Intelligence. Currently, he is a PhD student and an engineer at the Laboratory of Cognitive Dynamic Systems. His areas of interest include safe reinforcement learning and multimodal reinforcement learning. 
\end{IEEEbiography}

\begin{IEEEbiography}[{\includegraphics[width=1in,height=1.25in,clip,keepaspectratio]{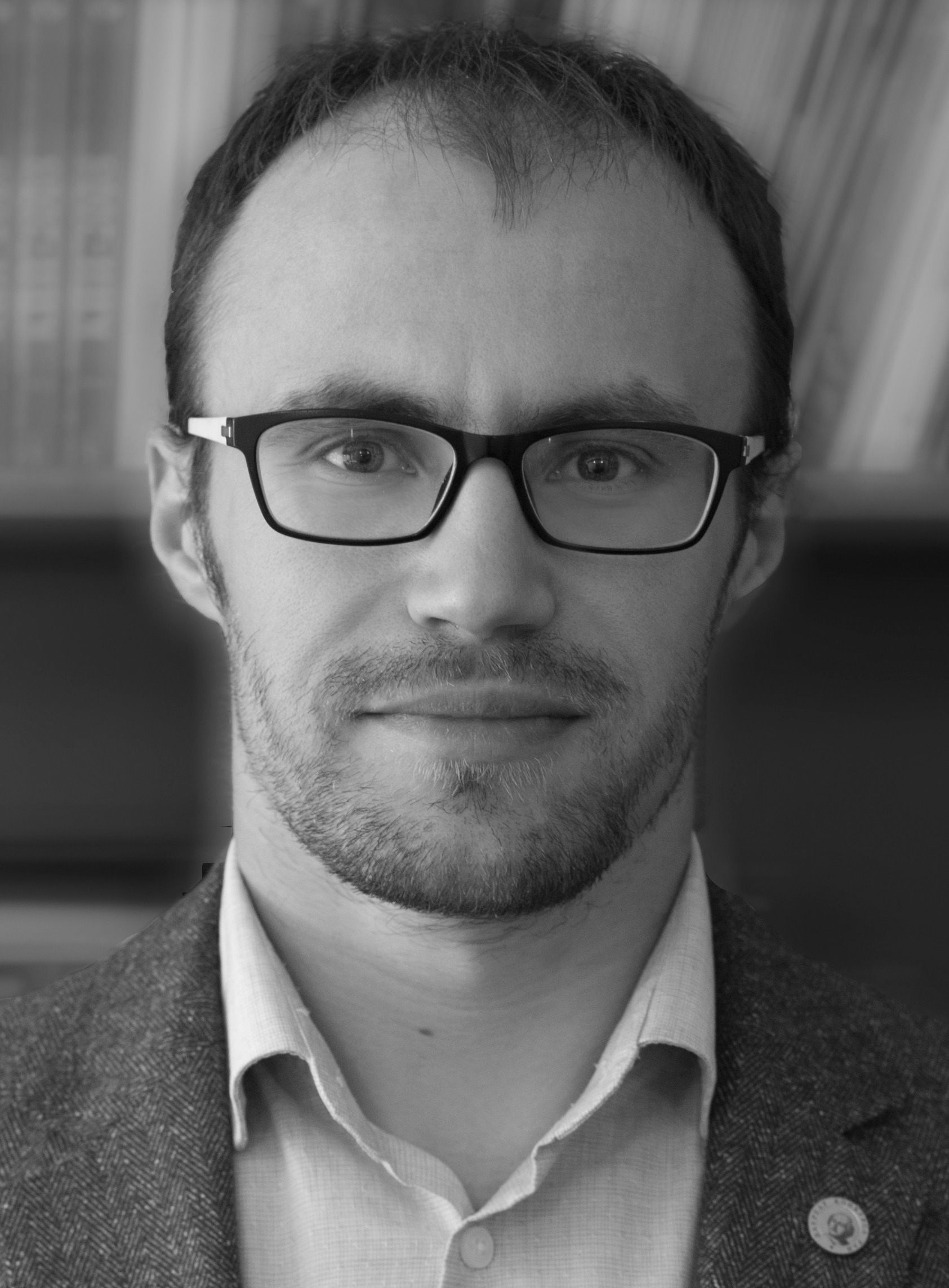}}]{Aleksandr I. Panov} earned an M.S. in Computer Science from the Moscow Institute of Physics and Technology, Moscow, Russia, 2011 and a Ph.D. in Theoretical Computer Science from the Institute for Systems Analysis, Moscow, Russia, in 2015.

Since 2010, he has been a research fellow with the Federal Research Center ``Computer Science and Control'' of the Russian Academy of Sciences. Since 2018, he has headed the Cognitive Dynamic System Laboratory at the Moscow Institute of Physics and Technology, Moscow, Russia. He authored three books and more than 100 research papers. In 2021, he joined the research group on Neurosymbolic Integration at the Artificial Intelligence Research Institute. His academic focus areas include behavior planning, reinforcement learning, embodied AI, and cognitive robotics. 
\end{IEEEbiography}

\begin{IEEEbiography}[{\includegraphics[height=1.3in,clip,keepaspectratio]{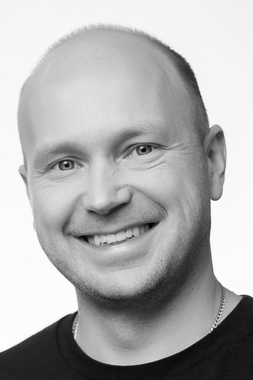}}]{Konstantin Yakovlev}
received the Ph.D. degree in computer science from the Institute for Systems Analysis, Russian Academy of Sciences, Moscow, Russia, in 2010.

He is currently a Leading Researcher at the Federal Research Center "Computer Science and Control" of the Russian Academy of Sciences, and is also affiliated with AIRI, MIPT and HSE University. His research interests include heuristic search, single and multi-agent pathfinding, motion planning, multi-agent systems, and robotics.
\end{IEEEbiography}

\EOD

\end{document}